# Synth4Seg - Learning Defect Data Synthesis for Defect Segmentation using Bi-level Optimization

Shancong Mou, Raviteja Vemulapalli, Shiyu Li, Yuxuan Liu, C Thomas, Meng Cao, Haoping Bai, Oncel Tuzel, Ping Huang, Jiulong Shan and Jianjun Shi

*Abstract*—Defect segmentation is crucial for quality control in advanced manufacturing, yet data scarcity poses challenges for state-of-the-art supervised deep learning. Synthetic defect data generation is a popular approach for mitigating data challenges. However, many current methods simply generate defects following a fixed set of rules, which may not directly relate to downstream task performance. This can lead to suboptimal performance and may even hinder the downstream task. To solve this problem, we leverage a novel bi-level optimization-based synthetic defect data generation framework. We use an online synthetic defect generation module grounded in the commonly-used Cut&Paste framework, and adopt an efficient gradient-based optimization algorithm to solve the bi-level optimization problem. We achieve simultaneous training of the defect segmentation network, and learn various parameters of the data synthesis module by maximizing the validation performance of the trained defect segmentation network. Our experimental results on benchmark datasets under limited data settings show that the proposed bi-level optimization method can be used for learning the most effective locations for pasting synthetic defects thereby improving the segmentation performance by up to 18.3% when compared to pasting defects at random locations. We also demonstrate up to 2.6% performance gain by learning the importance weights for different augmentation-specific defect data sources when compared to giving equal importance to all the data sources.

*Note to Practitioners*—Data limitations have long been a significant challenge for defect detection in the manufacturing industry. Consequently, the generation of synthetic defect samples is a crucial task. In our research, we developed an easy-to-implement, end-to-end, online differentiable Cut&Paste module that can be integrated into any image segmentation-based defect detection framework. This module can generate synthetic defect samples in real-time during the training of deep neural networks. Additionally, to ensure the quality of the generated synthetic defect samples, we introduced a bi-level optimization method that allows for controllable defect generation. This method is guided by validation set performance to enhance task-specific outcomes. Currently, our proposed method can not only learn optimal defect generation strategies but also identify critical areas on product surfaces for generating synthetic defects.

Shancong Mou is with the Department of Industrial and Systems Engineering, University of Minnesota, Minneapolis, MN, 55455. E-mail: mou00006@umn.edu.

Jianjun Shi is with the H. Milton Stewart School of Industrial and Systems Engineering, Georgia Institute of Technology, Atlanta, GA, 30332. E-mail:jianjun.shi@isye.gatech.edu.

Raviteja Vemulapalli, Shiyu Li, Yuxuan Liu, C Thomas, Meng Cao, Haoping Bai, Oncel Tuzel, Ping Huang, Jiulong Shan are with Apple Company, Cupertino, CA 95014.

E-mail: /r_vemulapalli, shiyu_li, yliu, c.thomas, mengcao, haoping_bai, ctuzel, huang_ping, jlshan/@apple.com.

## I. Introduction

Defect segmentation is an important component of quality control in modern manufacturing processes [1], [38]. With input images captured by industrial cameras, supervised segmentation approaches are commonly used to identify defective regions on various product surfaces. However, the lack of defect data is one of the biggest challenges in training a defect segmentation model. Although a great number of images are captured on the production lines, almost all of them are normal images without any defects because of the high yield rate of modern manufacturing pipelines.

One way to address the limited data problem is to generate synthetic defect data. Specifically, one can follow the popular Cut&Paste approach of [12] and generate new defective images by extracting the defect foregrounds from the limited training set, and pasting them at random locations on normal background images. To improve the synthetic data diversity, the cropped defects can be augmented using various random photometric and geometric transformations [32]. However, in real-world scenarios, there are different constraints on defect locations and textures for different products. As a result, some specific data augmentation methods and defect pasting locations may be more helpful than others to improve the defect segmentation performance for a product. For example, if defects can only happen on one particular region of the product, pasting synthetic defects on this specific region can speed up the model convergence. If all defect textures follow the same orientation due to the nature of the product surface or the manufacturing process, avoiding rotation augmentation may help reduce false positives. Hence, it is important to be able to learn the product-dependent optimal data augmentation strategies and effective pasting locations in an automated way.

To address this problem, we propose a bi-level optimization-based method that learns the data synthesis strategy in an online fashion, concurrent with the training process of the defect segmentation network. Concretely, we introduce a bi-level optimization [41] formulation, where the lower-level optimization problem focuses on training the defect segmentation network using the generated synthetic data, and the upper-level optimization problem focuses on minimizing the validation loss with respect to the learnable parameters of the data synthesis process. To learn which augmentations are important, we introduce an importance weight for each augmentation into our bi-level optimization formulation. Similarly, to learn effective pasting locations, we introduce a pasting location prediction network into our data synthesis process and optimize this network in our bi-level optimization framework (see Fig. 1). To solve the bi-level optimization problem efficiently, we use the gradient-based hyperparameter optimization approach of [27].

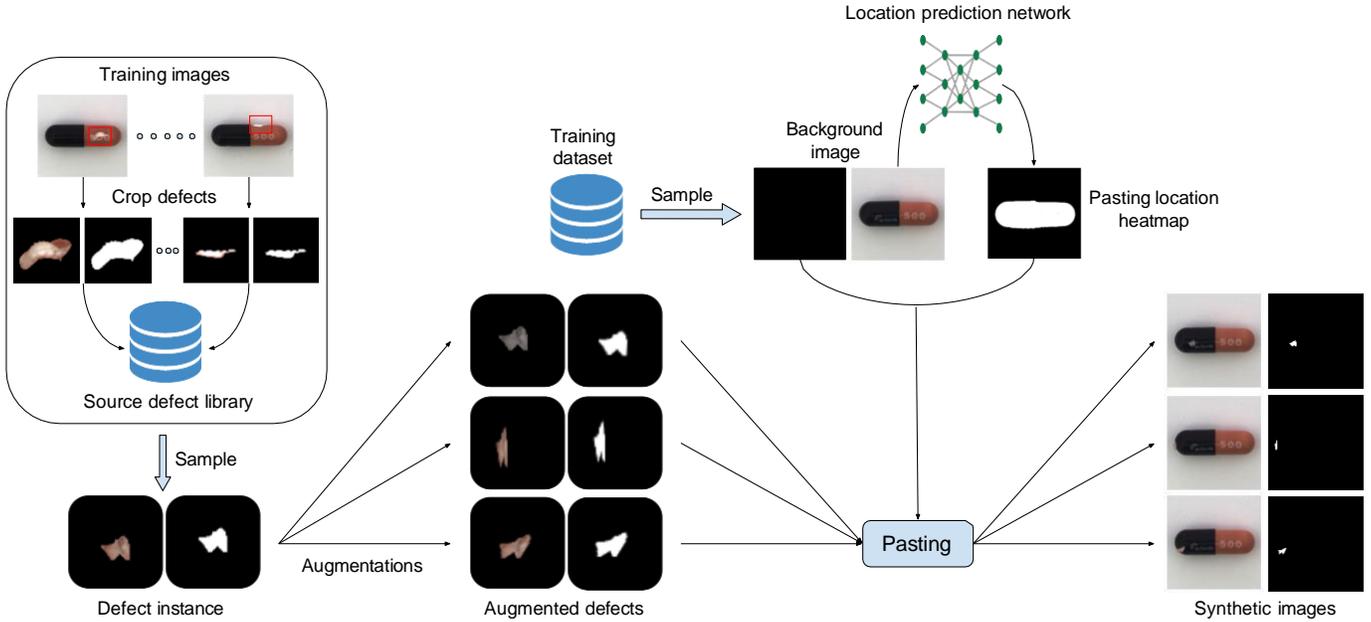

Fig. 1. Cut&Paste framework-based data synthesis module: First, a source defect library is created by cropping all the defect instances in the training set. During data synthesis, defect instances are randomly sampled from the source defect library and augmented using various image augmentation operations. These augmented defects are then pasted on randomly sampled training images by choosing the pasting locations based on a location heat map predicted by a pasting location prediction network.

We evaluate our method on MVTec [2], a benchmark dataset for industrial defect segmentation. When using multiple data augmentations, learning importance weights for these augmentations using our approach improves the defect segmentation performance by up to 2.6% when compared to training the segmentation network giving equal weight to all the augmentations. By learning a location prediction network to identify effective pasting locations, we observe gains of up to 18.3% in defect segmentation performance when compared to pasting defects randomly across the entire image.

### 1) Major contributions:

- We introduce a Cut&Paste framework-based defect data generation module that uses a local prediction network to identify the most effective defect pasting locations.
- We propose and implement a bi-level optimization framework that learns synthetic defect generation strategies for defect segmentation. We show how this bi-level optimization framework can be used to learn importance weights of various data augmentations and target image-dependent defect pasting locations.
- We demonstrate the effectiveness of the proposed bi-level optimization approach by conducting experiments on the MVTec benchmark dataset.

## II. RELATED WORK

Synthetic defect generation methods are crucial in the manufacturing defect segmentation setting, aimed at providing more diverse training images with defects. In this section, we provide a comprehensive review of the synthetic defect generation techniques.

**Data augmentation.** Data augmentation has emerged as a critical tool for enhancing data efficiency in the general computer vision domain tasks, which can be used in manufacturing defect segmentation. Among them, general-purpose data augmentation methods primarily focus on image-level augmentation, including photometric and geometric augmentation, with manually designed strategies [23], [39], [43]. These techniques have become standard preprocessing operations for defect segmentation. Importantly, general-purpose data augmentation retains the pixel-wise segmentation map. In recent years, the focus has shifted towards automatically determining the optimal set of augmentation parameters. Techniques such as AutoAugment [7], Fast AutoAugment [24], Population-based Augmentation [21], RandAugment [8] aim to find the best augmentation policy parameters through methods like grid search, heuristics, reinforcement learning, or bi-level optimization. Additionally, methods like Adversarial AutoAugment [49], focus on finding adversarial augmentation policies to generate challenging examples for target networks. While augmentation-learning methods can be applied to various tasks by adjusting the loss function, most are primarily focused on general classification problems. Improving policy search to be class-dependent [5] and learning augmentation based on individual samples [50] can enhance performance. Despite successes in learning-based approaches, Trivial Augment [32] stands out as non-learning-based approach that achieves superior performance.

**Generative model-based defect generation.** Generative models, such as Generative Adversarial Networks (GANs) [17] and diffusion models [35] are also popular methods in gener-ating synthetic defects. The main idea is to train a generative model on the training dataset and generate synthetic defect samples from the learned manifold. Representative methods include Defect-GAN [48], Defect Transfer GAN [45], and others [26], [46] (detailed review can be found in [31]). However, challenges exist, such as the quality of trained generative models on a small number of defect images and the lack of new information introduced in the synthetic data generation process. Another category relies on diffusion models pretrained on large-scale natural image datasets, such as Stable Diffusion [6], which is then fine-tuned on the defect dataset. While new information is introduced in this process, questions remain regarding its effectiveness, especially given the domain gap between natural and industrial images.

**Cut&Paste framework-based defect generation.** Cut&Paste [12] based strategies are specifically tailored for object detection or segmentation tasks. These methods not only apply image-level augmentations but also modify objects within images and relocate them. In the meanwhile, subsequent studies have investigated the impact of the object location [11], [16] and other augmentation parameters [44] on augmentation results. Those methods can be directly applied to synthetic defect generation in manufacturing applications. For example, Xu et al. [47] used the Cut&Paste strategy to generate synthetic defects for the inspection of lithium battery sealing nails; Pei et al. [33] designed a self-supervised anomaly detection based on Cut&Paste-generated synthetic defects; Ha et al. [18] further improved Cut&Paste to scaled CutPaste and FaultPaste for generating synthetic fault samples for planetary gearbox fault diagnosis.

In general, the Cut&Paste method comprises three steps: cutting out the defect from the source dataset, conducting augmentation, and pasting it back onto the target background image. These three steps make it a widely adopted and popular framework for synthetic defect generation in defect segmentation tasks. The two previously mentioned categories of synthetic defect generation methods can indeed be incorporated into the Cut&Paste framework. For example, data augmentation can be utilized to augment the cropped defects from the defect library, while the generative model-based approaches can be employed to generate additional defects, thus enlarging the defect library.

**Physical simulation and graphics-based defect generation.** Another category of synthetic defect generation algorithms is based on physical simulation and computer graphics [3], [15], [30], [37]. The main procedure involves first building the object of interest in simulation/rendering software, such as Blender [20], and then generating synthetic defects on the surface of the product. Finally, photorealistic synthetic inspection images are generated through computer graphics. Moonen et al. [30] proposed a Modular Toolkit for GPU-Accelerated photorealistic synthetic data generation for the manufacturing industry. Schmedemann et al. [4], [15], [37] developed a synthetic defect data generation pipeline based on procedural methods. Boikov et al. [3] generated synthetic defects on steel surfaces for defect inspection. Singh et al. [29], [40] combined finite element simulation and computer graphics to generate synthetic defects on manufactured products.

These approaches usually rely on prior knowledge of the defect type and can be labor-intensive and time-consuming.

**Learning-based defect generation and bi-level optimization.** Although learning data augmentation strategies based on validation performance has been extensively explored in the data augmentation domain [7], [8], [24], there are few works on learning synthetic defect generation strategies. In this paper, we propose a bi-level optimization framework for learning the optimal parameters in a synthetic defect generation pipeline for defect segmentation. One major challenge lies in the scale of such a bi-level optimization problem. For example, the defect location mask mentioned in the previous section is a high-dimensional upper-level decision variable. To learn such a high-dimensional upper-level decision variable, conventional grid search, heuristics, or Bayesian optimization may not be feasible. Therefore, gradient-based bi-level optimization algorithms [10], [13], [14], [25], [27], [28] are needed. These methods have been well developed in hyperparameter optimization of neural networks. In this paper, we adopt the inexpensive gradient-based hyperparameter optimization algorithm developed by [27], which combines the implicit function theorem (IFT) with efficient inverse Hessian approximations to handle millions of upper-level decision variables.

## III. APPROACH

In this section, we present our bi-level optimization-based approach for learning synthetic defect data generation to improve defect segmentation. First, we introduce our Cut&Paste framework-based data synthesis module in Sec. III-A. Then, we present our bi-level optimization framework for jointly optimizing the defect segmentation network and the data synthesis parameters in Sec. III-B.

### A. Defect Data Synthesis using Cut&Paste Module

The Cut&Paste approach [12] involves cutting objects of interest from a source dataset, applying augmentation operations to the cropped object instances, and then pasting them on images from the target dataset. This process generates new synthetic data samples with pixel-wise object segmentation masks. In this work, we first cut all the defect instances in the training dataset $\mathcal{T}$ to form a source defect library $\mathcal{D} = \{(D_i, M_i)\}_{i=1}^K$ in which each defect instance consists of a foreground texture image $D_i$ and a binary mask $M_i$ indicating the defect region (see Fig. 1 top-left). Then, we use the Cut&Paste approach to generate synthetic data on the fly while training the defect segmentation network by pasting randomly sampled defect instances on randomly sampled training images.

Specifically, in each training step, a set of $B$ defect instances $\{(D_i, M_i)\}_{i=1}^B$ are randomly sampled from the defect library $\mathcal{D}$. Then, $J$ data augmentation operations $\{\mathcal{A}_j\}_{j=1}^J$ are applied independently to each defect to generate $J$ batches of augmented defect instances $\mathcal{D}_j = \{(D_i^j, M_i^j)\}_{i=1}^B$, $j \in \{1, \ldots, J\}$. Next, $B$ labeled images $\{(X_i, Y_i)\}_{i=1}^B$ are randomly sampled from the training dataset $\mathcal{T}$ to be used as target images for pasting the defect instances. Here, $X_i$ represents an image and $Y_i$ represents the corresponding pixel-wise binary defect segmentation mask. If a sampled target image $X_i$ is defect-free, then $Y_i = 0$ at every pixel. The augmented defect instances are then pasted on these target images resulting in $J$ synthetic data batches $\mathcal{B}_j = \{(\tilde{X}_i^j, \tilde{Y}_i^j)\}_{i=1}^B$, $j \in \{1, \ldots, J\}$, which are used for training the defect segmentation network. Fig. 1 illustrates our data synthesis module using a single defect instance and a single background image.

*1) Pasting location.:* To paste a defect instance on a target image, we need to first choose a target location. Let $Z_i$ denote a binary mask that specifies the locations in the target image $X_i$ that are suitable for pasting defects. In our data synthesis module, the location for pasting a defect is uniformly sampled from all suitable locations indicated by $Z_i$. By default, we assume that the entire target image is suitable, i.e., $Z_i = 1$ at every pixel. While this assumption is valid for target images where the product of interest occupies the entire image, this is not optimal for target images where the product of interest only covers a subregion of the image. In such cases, we aim to learn a suitable location map $Z_i$ as a function of the target image $X_i$. To facilitate this, we model $Z_i$ as the output of a location prediction neural network $g_\theta$ that takes the target image $X_i$ as input, i.e., $Z_i = g_\theta(X_i)$.

*2) Pasting operation.:* For pasting a defect on a target image, we simply replace the content of the target image with the defect foreground content. Let $\Omega$ denote the region in the target image $X$ where a defect $D$ is being pasted. The resulting synthetic data sample $(\tilde{X}, \tilde{Y})$ is given by

$$\tilde{X}(p) = X(p), \tilde{Y}(p) = Y(p) \text{ for } p \notin \Omega, \quad (1)$$
$$\tilde{X}(p) = D(p), \tilde{Y}(p) = 1, \text{ for } p \in \Omega,$$

where $Y$ is the defect mask of the target image $X$.

As an alternative to this direct pasting, we also considered Poisson image blending [34], which is a more sophisticated pasting operation that ensures a smooth transition on the blended region boundary. But, using Poisson blending instead of direct pasting did not produce any noticeable improvement in the defect segmentation performance in our early experiments. Hence, we use the simple approach of direct pasting in this work. Algorithm 1 summarizes all the steps in our data synthesis module.

### B. Optimization of Segmentation Network and Synthesis Module

Given our synthesis module, we can generate synthetic data batches $\{\mathcal{B}_j\}_{j=1}^J$ on the fly, and train the defect segmentation network $f_w$ using

$$w^* = \arg\min_w \sum_{j=1}^J \eta_j \mathbb{E}_{(\tilde{X}, \tilde{Y}) \in \mathcal{B}_j} \left[ \mathcal{L}\left(f_w(\tilde{X}), \tilde{Y}\right) \right], \quad (2)$$

where $\mathcal{L}$ is the training loss function and $\boldsymbol{\eta} = \{\eta_j \geq 0\}_{j=1}^J$ are the importance weights for different synthetic data sources $\{\mathcal{B}_j\}_{j=1}^J$.

*1) Learning the importance of different synthetic data sources:* A wide variety of data augmentations [32] exist that can be used for data synthesis in our Cut&Paste module, and each augmentation operation can be interpreted as one

---

**Algorithm 1:** Defect data synthesis using Cut&Paste module

**Inputs** - Training dataset $\mathcal{T}$, Defect library $\mathcal{D}$, Direct pasting operation $\oplus_l$ where $l$ denotes the pasting location, Augmentation operations $\{\mathcal{A}_j\}_{j=1}^J$, Optional location prediction network $g_\theta$.

Sample $B$ defect instances $\{(D_i, M_i)\}_{i=1}^B$ from the defect library $\mathcal{D}$.

*# Generate augmented defects.*
**for** $j = 1, \ldots, J$ **do**
    **for** $i = 1, \ldots, B$ **do**
        $(D_i^j, M_i^j) = \mathcal{A}_j((D_i, M_i))$.
    **end**
**end**

Sample $B$ labeled target images $\{(X_i, Y_i)\}_{i=1}^B$ from the training dataset $\mathcal{T}$.

*# Compute location maps for pasting defects.*
**for** $i = 1, \ldots, B$ **do**
    **if** *using location prediction network* **then**
        $Z_i = g_\theta(X_i)$
    **else**
        $Z_i = 1$ at every pixel.
    **end**
**end**

*# Paste augmented defects on target images.*
**for** $j = 1, \ldots, J$ **do**
    **for** $i = 1, \ldots, B$ **do**
        Randomly sample a target location $l$ based on $Z_i$.
        $(\tilde{X}_i^j, \tilde{Y}_i^j) = (D_i^j, M_i^j) \oplus_l (X_i, Y_i)$
    **end**
**end**
**Output:** Synthetic data batches
$\mathcal{B}_j = \{(\tilde{X}_i^j, \tilde{Y}_i^j)\}_{i=1}^B$, $j \in \{1, \ldots, J\}$

---

synthetic data source. Depending on the target product and defects, some of the the data sources could be more useful than the others. Hence, we propose to learn their importance weights $\boldsymbol{\eta}$ by solving the following bi-level optimization problem:

$$\min_{\boldsymbol{\eta}} \ \mathcal{L}_V, \ \mathcal{L}_V = \mathbb{E}_{(X,Y) \in \mathcal{V}} \left[ \mathcal{L}(f_{w^*}(X), Y) \right],$$
$$w^* = \arg\min_w \ \mathcal{L}_T, \quad (3)$$
$$\mathcal{L}_T = \sum_{j=1}^J \eta_j \mathbb{E}_{(\tilde{X}, \tilde{Y}) \in \mathcal{B}_j} \left[ \mathcal{L}\left(f_w(\tilde{X}), \tilde{Y}\right) \right],$$

where $\mathcal{V}$ denotes a held-out validation dataset that is different from the training dataset $\mathcal{T}$. The lower-level optimization problem in (3) solves for the optimal defect segmentation network $f_w$ for given importance weights $\boldsymbol{\eta}$ and the upper-level optimization problem solves for optimal $\boldsymbol{\eta}$ that minimizes the validation loss of the defect segmentation network.

*2) Learning the most effective locations for pasting defects:* One can consider the entire target image as a valid region for

pasting defects and randomly choose any location for pasting. However, depending on the product, adding synthetic defects at certain locations (that are more prone to defects or make defects harder to distinguish) could be more effective than other locations for training the segmentation network. Hence, we propose to learn the set of effective pasting locations as a function of the target image. To facilitate this, we use a location prediction network $g_\theta$ as described in Sec. III-A.

Intuitively, when $g_\theta$ is trained well, the synthetic images generated by pasting defects on a target image $X$ at locations where $g_\theta(X)$ is high should be given more weight while training the defect segmentation network when compared to the synthetic images generated by pasting defects on $X$ at locations where $g_\theta(X)$ is low. To capture this, we introduce a sample weight $\alpha(\tilde{X})$ into the optimization problem corresponding to segmentation network training:

$$\arg\min_w \sum_{j=1}^{J} \eta_j \mathbb{E}_{(\tilde{X},\tilde{Y}) \in \mathcal{B}_j} \left[ \alpha(\tilde{X}) \mathcal{L} \left( f_w(\tilde{X}), \tilde{Y} \right) \right],$$
$$\alpha(\tilde{X}) = \frac{\|g_\theta(X) \odot M\|_1}{\|M\|_1}, \tag{4}$$

where $M$ is a binary mask indicating the region in $X$ where a defect has been pasted to synthesize $\tilde{X}$, and $\odot$ denotes element-wise multiplication. If the location prediction network output $g_\theta(X)$ is high in the pasted defect region, then the corresponding synthetic sample $\tilde{X}$ gets a higher weight $\alpha(\tilde{X})$. Using (4) as the lower-level optimization problem, we propose the following bi-level optimization problem for jointly training the location prediction network $g_\theta$ and the defect segmentation network $f_w$:

$$\min_\theta \ \mathcal{L}_V, \ \mathcal{L}_V = \mathbb{E}_{(X,Y) \in \mathcal{V}} \left[ \mathcal{L}(f_{w^\star}(X), Y) \right]$$
$$+ \gamma \mathbb{E}_{X \in \mathcal{T}} [\|g_\theta(X)\|_1],$$
$$w^\star = \arg\min_w \mathcal{L}_T, \tag{5}$$
$$\mathcal{L}_T = \sum_{j=1}^{J} \eta_j \mathbb{E}_{(\tilde{X},\tilde{Y}) \in \mathcal{B}_j} \left[ \alpha(\tilde{X}) \mathcal{L} \left( f_w(\tilde{X}), \tilde{Y} \right) \right].$$

Here, we use an additional sparsity inducing $\ell_1$ regularization loss on $g_\theta(X)$ to encourage the location prediction network to identify the most important locations for pasting synthetic defects.

*Remark 1.* While we focus on learning the importance of different synthetic data sources and effective image locations for pasting defects, the proposed bi-level optimization framework can easily be extended to optimize other components of our data synthesis module such as the parameters of the data augmentation operations.

*Remark 2.* Solving optimization problems (3) and (5) efficiently is non-trivial, especially for high-dimensional problems. We will introduce the adopted bi-level optimization method [27] in the next section.

## IV. A GRADIENT BASED BI-LEVEL OPTIMIZATION ALGORITHM FOR

To solve this bi-level optimization problem efficiently, we give an introduction of the gradient-based hyperparameter optimization approach from [27], as well as the detailed algorithm to solve the optimization problems (3) and (5).

Optimization problems (3) and (5) are instances of the following bi-level optimization problem:

$$\min_p \ \mathcal{L}_V(p, w^\star(p)),$$
$$w^\star = \arg\min_w \ \mathcal{L}_T(p, w),$$

where $p$ denotes the importance weights $\eta$ in the case of (3) and location prediction network parameters $\theta$ in the case of (5), and $w$ denotes the defect segmentation network parameters.

In order to use standard gradient-based optimization approaches for solving this bi-level optimization problem, we need to compute the hyper-gradient $\partial \mathcal{L}_V(p, w^\star(p))/\partial p$, which can be written as

$$\left. \frac{\partial \mathcal{L}_V(p, w^\star(p))}{\partial p} \right|_{p'} \tag{6}$$
$$= \left. \frac{\partial \mathcal{L}_V(p, w)}{\partial p} \right|_{p', w^\star(p')} + \left. \frac{\partial \mathcal{L}_V(p, w)}{\partial w} \right|_{p', w^\star(p')} \times \left. \frac{\partial w^\star(p)}{\partial p} \right|_{p'}.$$

The term $\partial w^\star(p)/\partial p$ is challenging to compute because it must account for how the optimal neural network weights change with respect to the hyperparameters $p$. Following [27], this term is approximated using the implicit function theorem (IFT), i.e.,

$$\left. \frac{\partial w^\star(p)}{\partial p} \right|_{p'}$$
$$= - \left[ \frac{\partial^2 \mathcal{L}_T(p, w)}{\partial w \partial w^\top} \right]^{-1} \left[ \frac{\partial^2 \mathcal{L}_T(p, w)}{\partial w \partial p^\top} \right] \Bigg|_{p', w^\star(p')}, \tag{7}$$

where the intractable inversion of Hessian $-\left[ \partial^2 \mathcal{L}_T / \partial w \partial w^\top \right]^{-1}$ is further approximated by a finite Neumann series as

$$\left[ \frac{\partial^2 \mathcal{L}_T(p, w)}{\partial w \partial w^\top} \right]^{-1} \approx \sum_{k=0}^{K} \left[ I - \frac{\partial^2 \mathcal{L}_T(p, w)}{\partial w \partial w^\top} \right]^k, \tag{8}$$

where $K$ is the number of terms used to achieve satisfactory approximation performance. For stability, we use the following implementation:

$$\left[ \frac{\partial^2 \mathcal{L}_T(p, w)}{\partial w \partial w^\top} \right]^{-1} = \alpha \left[ \alpha \frac{\partial^2 \mathcal{L}_T(p, w)}{\partial w \partial w^\top} \right]^{-1}$$
$$\approx \alpha \sum_{k=0}^{K} \left[ I - \alpha \frac{\partial^2 \mathcal{L}_T(p, w)}{\partial w \partial w^\top} \right]^k, \tag{9}$$

where $\alpha$ is the learning rate.

Finally, the hyper-gradient $\partial \mathcal{L}_V(p, w^\star(p))/\partial p$ can be calculated as

$$\left. \frac{\partial \mathcal{L}_V(p, w^\star(p))}{\partial p} \right|_{p'} = \left. \frac{\partial \mathcal{L}_V(p, w)}{\partial p} \right|_{p', w^\star(p')} \tag{10}$$
$$- \alpha \sum_{k=0}^{K} \frac{\partial \mathcal{L}_V(p, w)}{\partial w} \left[ I - \alpha \frac{\partial^2 \mathcal{L}_T(p, w)}{\partial w \partial w^\top} \right]^k \left[ \frac{\partial^2 \mathcal{L}_T(p, w)}{\partial w \partial p^\top} \right] \Bigg|_{p', w^\star(p')}.$$

Algorithm 2 presents the gradient-based optimization approach used for solving the above bi-level optimization problem.

---

**Algorithm 2: Bi-level optimization algorithm for learning $p$ and $w$.**

**Input:** Inner optimization problem optimization steps $N$, maximum iterations $N_{max}$, Truncation number of Neumann series $K$, Learning rate $\alpha$ for segmentation network parameters, Learning rate $\beta$ for data synthesis parameters, warm up steps $N_0$.

**Initialize** data synthesis parameters $p$ ($p$ is the importance weights $\eta$ in (3) and location prediction network weights $\theta$ in (5)) and segmentation network parameters $w$.

**for** $i = 1, \ldots, N_{max}$ **do**

    Execute Algorithm 1 in the main paper to generate synthetic defect image and annotation pairs $\{(\tilde{X}_i^j, \tilde{Y}_i^j)\}_{i=1}^B, j \in \{1, \ldots, J\}$ ;

    Gradient descent step of the neural network weights:

$$w \leftarrow w - \alpha \left( \frac{\partial \mathcal{L}_T}{\partial w} \right)$$

    **if** $(i \geq N_0)$ & $(i \mod N = 0)$ **then**

        Calculate $\partial \mathcal{L}_V(p, w)/\partial w$ and $\partial \mathcal{L}_V(p, w)/\partial p$ using validation dataset $\mathcal{V}$.

        Calculate Neumann series approximation of the Inverse-Hessian-vector product through the vector-Jacobian Products (VJP):

$$\alpha \frac{\partial \mathcal{L}_V}{\partial w} \sum_{k=0}^{K} \left[ I - \alpha \frac{\partial^2 \mathcal{L}_T}{\partial w \partial w^\top} \right]^k$$

        by the following algorithm:

        \*\*\*\*\*\*\*\*\*\*\*\*\*\*\*\*\*\*

        Initialize $\lambda = \frac{\partial \mathcal{L}_V}{\partial w}, v = \frac{\partial \mathcal{L}_V}{\partial w}$;

        **for** $k = 1, \ldots, K$ **do**

            $v \leftarrow v - \alpha v \frac{\partial^2 \mathcal{L}_T}{\partial w \partial w^\top}$

            $\lambda \leftarrow \lambda + v$

        **end**

        Return $\alpha \lambda$;

        \*\*\*\*\*\*\*\*\*\*\*\*\*\*\*\*\*\*

        Calculate:

$$\frac{\partial \mathcal{L}_V}{\partial p} = \frac{\partial \mathcal{L}_V(p, w)}{\partial p} - \alpha \lambda \frac{\partial^2 \mathcal{L}_T(p, w)}{\partial w \partial p^\top}$$

        Gradient descent step of augmentation policy parameters:

$$p \leftarrow p - \beta \left( \frac{\partial \mathcal{L}_V}{\partial p} \right)$$

    **end**

**end**

---

| Product | Training | | Validation | Test |
|---|---|---|---|---|
| | Defective | Clean | Defective | Defective |
| Carpet | 43 | 281 | 26 | 23 |
| Grid | 28 | 265 | 16 | 16 |
| Leather | 45 | 246 | 26 | 24 |
| Wood | 30 | 248 | 18 | 15 |
| Tile | 39 | 231 | 24 | 21 |
| Screw | 59 | 321 | 34 | 29 |
| Capsule | 54 | 220 | 31 | 27 |

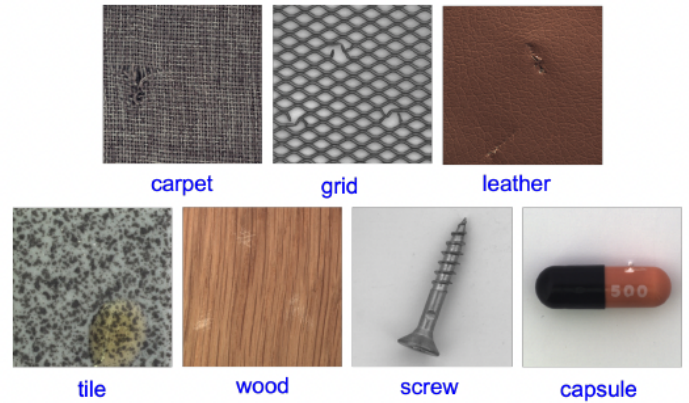

Fig. 2. Example images of seven products from the MVTec [2] dataset.

## V. EXPERIMENTS

In this section, we present our experimental results that demonstrate the effectiveness of the proposed bi-level optimization framework in terms of learning the importance weights of different synthetic data sources and the most effective target image locations for pasting defects.

**Dataset.** We use the benchmark MVTec [2] anomaly detection dataset which provides high quality pixel-level defect annotations for a diverse set of defect types and products. Specifically, we use seven different products from this dataset and conduct separate experiments for each product. Fig. 2 shows example images of these products and Tab. I shows the number of training, validation and test images for each product. The training and validation splits are used for our bi-level optimization and the test split is used to report the final results.

**Network architectures.** We use a standard U-Net architecture [36] with a ResNet18 [19] encoder and a pixel-wise sigmoid classification layer for both defect segmentation and location prediction networks. The encoder is always initialized with a ResNet18 checkpoint pretrained on the ImageNet dataset [9].

**Loss function.** We use an equal-weighted average of binary cross-entropy loss and Dice loss [42] for training.

### A. Learning Data Source Importance Weights

We use four different random data augmentations, namely *photometric*, *rotation*, *shear* and *scale*, in our Cut&Paste synthesis framework resulting in four synthetic data sources, namely *Cut&Paste-Photometric*, *Cut&Paste-Rotation*, *Cut&Paste-Shear* and *Cut&Paste-Scale*. The photometric augmentation involves adjusting brightness, contrast and saturation. Tab. II presents the augmentation strength ranges used for each of these augmentation operations. In addition to these, we also consider a fifth synthetic data source, referred to as *TrivialAug-Global*, which applies the popular TrivialAugment [32] operation directly on the original training images (no Cut&Paste). Apart from these five synthetic data sources, we also use a sixth data source, referred to as *Defect-free*, consisting of real images without any defects.

TABLE II
AUGMENTATION OPERATIONS AND THEIR STRENGTH RANGES ADOPTED FROM [32].

| Augmentation | Photometric | | | Shear | | Rotation | Scale |
|---|---|---|---|---|---|---|---|
| | Brightness | Contrast | Saturation | Shear_x | Shear_y | | |
| Strength range | (0.1, 1.9) | (0.1, 1.9) | (0.1, 1.9) | (−0.3, 0.3) | (−0.3, 0.3) | (−30°, 30°) | (0, 2) |

TABLE III
DEFECT IoU (MEAN/STD) ON THE TEST SET. LEARNING IMPORTANCE WEIGHTS FOR DIFFERENT DATA SOURCES PERFORMS BETTER THAN GIVING EQUAL IMPORTANCE TO ALL DATA SOURCES.

| Synthetic data source | Carpet | Grid | Leather | Wood | Tile |
|---|---|---|---|---|---|
| TrivialAug-Global | 0.573/0.007 | 0.519/0.015 | 0.626/0.019 | 0.598/0.016 | 0.720/0.007 |
| Multiple (equal weights) | 0.643/0.001 | 0.570/0.002 | 0.705/0.002 | 0.702/0.016 | 0.794/0.013 |
| Multiple (learned weights) | **0.662**/0.004 | **0.576**/0.004 | **0.719**/0.007 | **0.706**/0.030 | **0.820**/0.009 |

We jointly optimize the defect segmentation network parameters $w$ and the data source importance weights $\eta$ by solving the bi-level optimization problem (3). We initialize all the importance weights with a value of one and train for 150 epochs using the Adam optimizer [22] with an initial learning rate of $2.5e^{-4}$ which is reduced by half every 30 epochs. We sample two defect instances and two training images in each training step ($B = 2$). We train the segmentation network for the first 30 epochs using the default equal data source importance weights, and update both the data source importance weights and segmentation network parameters for the remaining 120 epochs. We fix one of the importance weights to be one and learn only the remaining weights such that they capture the relative importance of different data sources.

We experiment with five products from the MVTec dataset, namely Carpet, Grid, Leather, Wood and Tile. Tab. III compares the defect segmentation performance achieved by our bi-level optimization approach with the performance achieved by training the segmentation network giving equal weight to all six data sources, i.e., $\eta_j = 1, \forall j$. We use the standard defect intersection-over-union (IoU) metric computed on a held-out test set as our evaluation metric. We ran each experiment three times and report the mean and standard deviation. Learning the importance weights clearly performs better than giving equal importance to all the data sources. Specifically, we observe 2.6%, 1.9% and 1.4% improvement in the case of Tile, Carpet and Leather, respectively. The proposed approach also significantly outperforms the TrivialAugment-Global baseline that only uses synthetic data generated by applying data augmentations to the entire image.

Fig. 3 shows the importance weights of different data sources learned by our bi-level optimization for the Carpet and Leather datasets. Weights $\eta_1$ to $\eta_6$ correspond to TrivialAug-Global, Cut&Paste-Photometric, Cut&Paste-Rotation, Cut&Paste-Shear, Cut&Paste-Scale and Defect-free data sources, respectively. In the case of Carpet dataset, the learned weights suggest that Defect-free and all Cut&Paste data sources and much more important than the TrivialAug-Global data source. Among the Cut&Paste data sources, Cut&Paste-Shear is the most important one and Cut&Paste-Scale is the least important one. In the case of Leather dataset, Defect-free, Cut&Paste-Photometric and Cut&Paste-Shear data sources are much more important than TrivialAug-Global, Cut&Paste-Rotation and Cut&Paste-Scale data sources. Learned weights for the remaining three product datasets can be found in the supplementary material.

### B. Learning Effective Pasting Locations

In the case of Carpet, Grid, Leather, Wood and Tile, the product covers the entire image making the whole image equally good for pasting defects. Different from these products, in the case of Screw and Capsule, the product only covers a subregion of the image. In such cases, randomly pasting defects everywhere in the image may not be optimal. Hence, we use a location prediction network to identify the most effective pasting locations when experimenting with Screw and Capsule datasets. Our main focus here is on demonstrating the ability of our bi-level optimization framework to learn effective pasting locations. So, instead of defining multiple data sources each corresponding to a different augmentation as in Sec. V-A, we combine multiple augmentations into a single data source ($J = 1$ in (5)) and focus on learning the location prediction network in the bi-level optimization. Specifically, we generate synthetic data by applying every augmentation from Tab. II with a probability of 0.5 to each defect instance within our Cut&Paste framework.

We train the location prediction network and the defect segmentation network jointly by solving the bi-level optimiza-tion problem (5). We train for a total of 60 epochs using the Adam optimizer. For the first 30 epochs, we only update the defect segmentation network by considering the entire target image as a valid region for pasting defects. After that, we simultaneously update both the location prediction and defect segmentation networks while using the output of the location prediction network to sample pasting locations. Specifically, in each iteration, we uniformly sample the pasting location from the set of locations where the output of the current location prediction network is above a certain threshold. [1] We use an initial learning rate of $2.5e^{-4}$ for training the segmentation network and reduce this by half after 30 epochs. We use a learning rate of $1e^{-4}$ for training the location prediction network. We sample two defect instances and two training images in each training step ($B = 2$). The weight $\gamma$ for the $\ell_1$ regularization term in (5) is set to $1e^{-4}$. For reporting the final results on the held-out test set, we use the segmentation network checkpoint that corresponds to the best performance on the validation dataset.

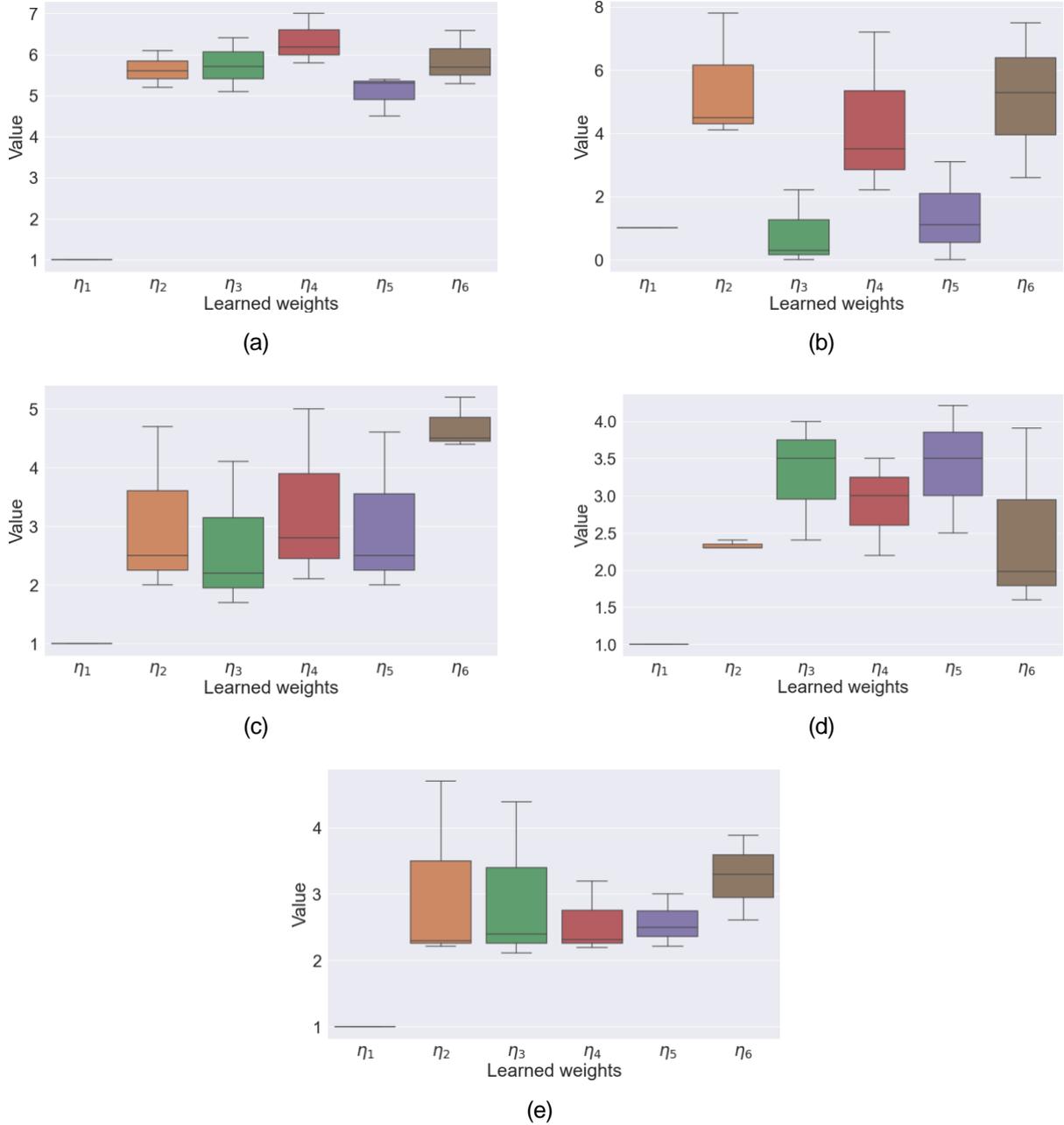

Fig. 3. Learned importance weights for various data sources. Weights $\eta_1$ to $\eta_6$ correspond to TrivialAug-Global, Cut&Paste-Photometric, Cut&Paste-Rotation, Cut&Paste-Shear, Cut&Paste-Scale and Defect-free data sources, respectively. The box plots are based on data from three random trials. (a) Carpet; (b) Leather; (c) Grid; (d) Wood; (e) Tile.

Tab. IV compares the segmentation performance achieved by our bi-level optimization approach with the performance of a network trained by pasting defects at random locations.

---

[1] We use a threshold of 0.6 for Capsule and 0.7 for Screw.

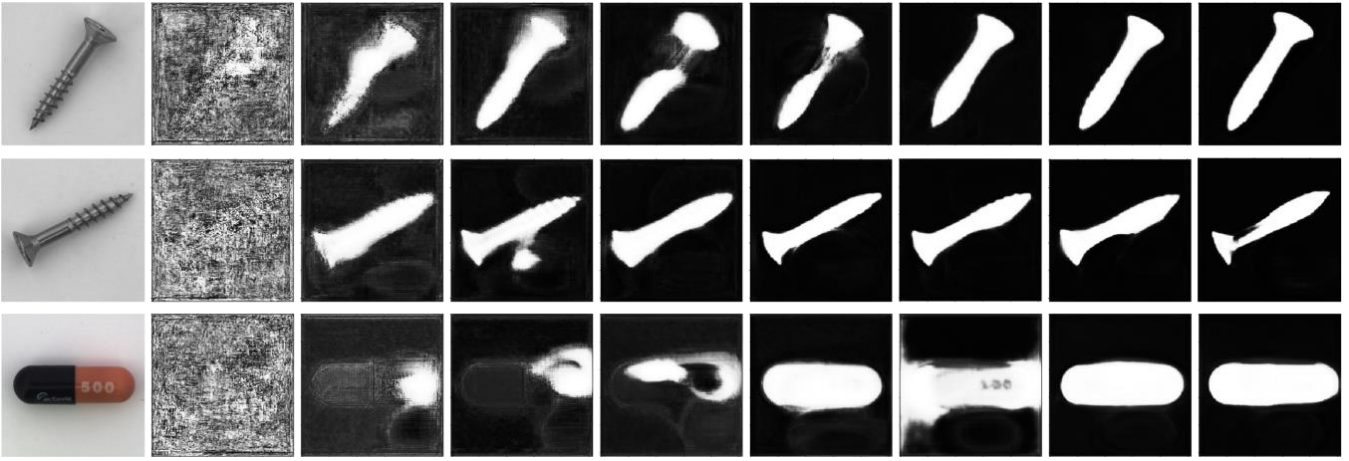

Fig. 4. Pasting location heat maps produced by the location prediction network at different points of time during training (epochs {30, 33, ..., 48, 51}). The location prediction network progressively learns to identify the product foreground region as the most important region for pasting defects without having any prior knowledge of the product location in the image.

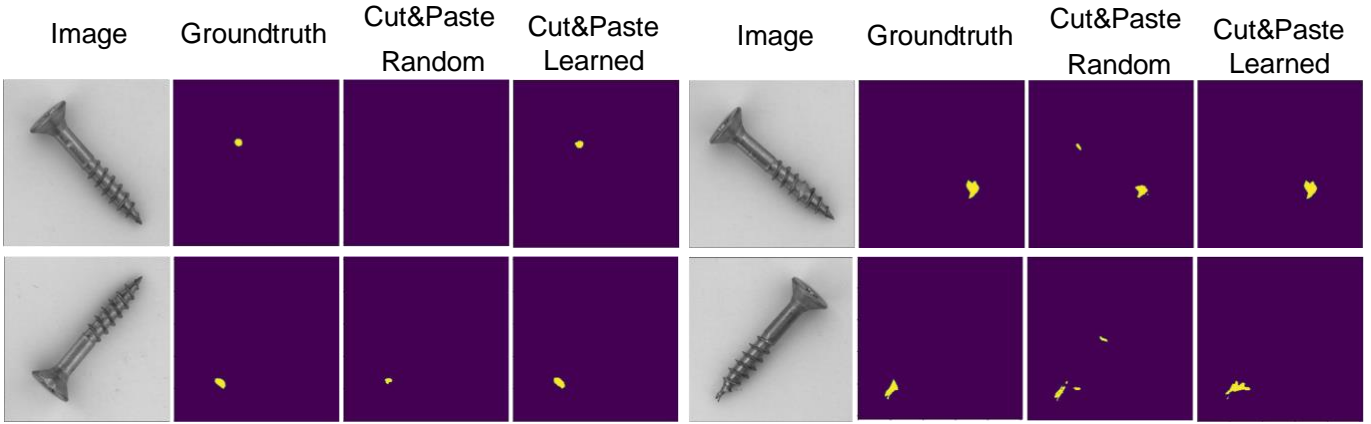

Fig. 5. Segmentation results on test images from the Screw dataset. Cut&Paste Random refers to pasting defects at random locations and Cut&Paste-Learned refers to the proposed approach that pastes defects at learned locations.

TABLE IV

Defect IoU on the test set. Pasting defects at learned locations performs significantly better than pasting at random locations, and is even comparable to pasting using groundtruth product location.

| Synthesis approach | Screw | Capsule |
|---|---|---|
| TrivialAug-Global | 0.323 | 0.374 |
| Cut&Paste - Random locations | 0.301 | 0.418 |
| Cut&Paste - Learned locations | 0.484 | 0.540 |
| Cut&Paste - Groundtruth product location | 0.468 | 0.551 |

Pasting defects at learned locations performs significantly better than pasting at random locations. Specifically, we observe 18.3% improvement for the Screw dataset and 12.2% improvement for the Capsule dataset. We also significantly outperform the TrivialAugment-Global baseline that generates synthetic data by applying data augmentations to the entire image. In fact, the performance of the proposed approach is close to the performance one would obtain when the groundtruth product location is used for selecting the defect pasting locations.

Fig. 4 shows how the location prediction network progressively learns to identify the important pasting locations during our bi-level optimization. At the beginning, the output of the location prediction network is high all over the image and defects are pasted everywhere. As training progresses, the back-propagated gradient from the validation set is prompting the location prediction network to reduce its prediction values outside the product foreground regions since pasting defects on the background regions does not necessarily produce synthetic data that is useful for improving defect segmentation on product surfaces. Eventually, the network identifies the product foreground region as the most important region for pasting defects. Fig. 5 and Fig. 6 shows the defect segmentation results for few test images from the screw dataset. Training with pasting defects at random locations results in a segmentation network that either misses the defects or generates false positives. In comparison, the network trained with the proposed bi-level optimization approach (that learns effective pasting locations) produces much better results. See the supplementary material for additional visual results.

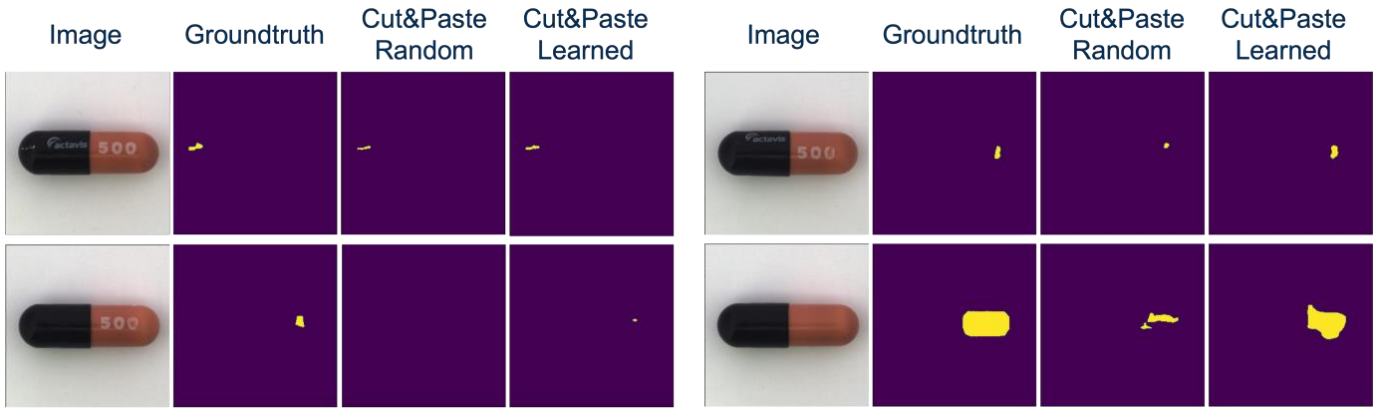

Fig. 6. Segmentation results on test images from the Screw dataset. Cut&Paste Random refers to pasting defects at random locations and Cut&Paste-Learned refers to the proposed approach that pastes defects at learned locations.

## C. Learning Augmentation Range Parameters

Apart from learning importance weights for various augmentations and effective pasting locations, we also explored learning the range parameters of the photometric and geometric augmentations using our bi-level optimization framework. Specifically, we set the importance weights for all augmentation-specific data sources to be 1 and considered the augmentation range parameters as learnable hyper-parameters. Unfortunately, we did not observe any noticeable performance gain by learning the range parameters when compared to using the default range values provided in Tab. II. One possible reason for this could be that the default range values are already good and not sensitive to small perturbations.

## VI. CONCLUSIONS

This paper introduces a Cut&Paste framework-based defect data synthesis module that uses a location prediction network for identifying the most effective regions in the target image for pasting defects, and a bi-level optimization framework to simultaneously optimize the defect segmentation network and data synthesis module parameters. The proposed bi-level optimization framework can be used to learn the relative importance of various data augmentation operations and target image-dependent defect pasting locations. Our experimental results on the benchmark MVTec dataset show that the proposed approach clearly outperforms the competing baselines. Specifically, we achieve significant performance improvements by learning where to paste.